\documentclass{article}

\usepackage[T1]{fontenc}

\usepackage[bitstream-charter]{mathdesign}
\usepackage{amsmath}
\usepackage[scaled=0.92]{PTSans}

\newtheorem{lemma}{Lemma}
\newtheorem{corollary}{Corollary}

\usepackage[
  paper  = letterpaper,
  left   = 1.5in,
  right  = 1.5in,
  top    = 1.0in,
  bottom = 1.0in,
  ]{geometry}
\usepackage{setspace}

\usepackage[usenames,dvipsnames]{xcolor}
\definecolor{shadecolor}{gray}{0.9}

\usepackage[final,expansion=alltext]{microtype}
\usepackage[english]{babel}
\usepackage[parfill]{parskip}
\usepackage{afterpage}
\usepackage{framed}
\usepackage{nicefrac}

\DeclareRobustCommand{\parhead}[1]{\textbf{#1}~}

\usepackage{graphicx}
\usepackage[labelfont=bf]{caption}
\usepackage[format=hang]{subcaption}

\usepackage{booktabs}

\usepackage{natbib}

\usepackage[algoruled]{algorithm2e}
\usepackage{listings}
\usepackage{fancyvrb}
\fvset{fontsize=\normalsize}

\usepackage[colorlinks,linktoc=all]{hyperref}
\usepackage[all]{hypcap}
\hypersetup{citecolor=Violet}
\hypersetup{linkcolor=black}
\hypersetup{urlcolor=MidnightBlue}

\usepackage[acronym,smallcaps,nowarn]{glossaries}
\glsdisablehyper
\makeglossaries

\newcommand{\red}[1]{\textcolor{BrickRed}{#1}}

\DeclareRobustCommand{\mb}[1]{\ensuremath{\boldsymbol{\mathbf{#1}}}}

\renewcommand{\mid}{\,\vert\,}
\newcommand{\prm}{\:;\:}

\newcommand{\mbx}{\mb{x}}

\newcommand{\mbtheta}{\mb{\theta}}

\newcommand{\mbphi}{\mb{\phi}}

\newcommand{\mbbeta}{\mb{\beta}}

\newcommand{\mbeta}{\mb{\eta}}
\newcommand{\mbmu}{\mb{\mu}}
\newcommand{\mbpi}{\mb{\pi}}

\newcommand{\mbxi}{\mb{\xi}}

\newcommand\dif{\mathop{}\!\mathrm{d}}

\newcommand{\E}{\mathbb{E}}
\newcommand{\V}{\mathbb{V}}

\newcommand{\cN}{\mathcal{N}}

\newcommand{\Gam}{\textrm{Gam}}

\newacronym{ELBO}{elbo}{evidence lower bound}
\newacronym{GMM}{gmm}{Gaussian mixture model}
\newacronym{KL}{kl}{Kullback-Leibler}
\newacronym{LDA}{lda}{latent Dirichlet allocation}
\newacronym{SVI}{svi}{stochastic variational inference}
\newacronym{MCMC}{mcmc}{Markov chain Monte Carlo}
\newacronym{ADVI}{advi}{automatic differentiation variational inference}

\newacronym[firstplural={posterior dispersion indices},
            \glsshortpluralkey={pdi}]
{PDI}{pdi}{posterior dispersion index}

\newacronym{WAPDI}{wapdi}{widely applicable posterior dispersion index}
\newacronym{WAIC}{waic}{widely applicable information criterion}

\newacronym{HPF}{hpf}{hierarchical Poisson factorization}

\newacronym{SNP}{snp}{single-nucleotide polymorphism}
 
\title{\textbf{Posterior Dispersion Indices}}

\author{
  Alp Kucukelbir\\
  Columbia University\\ \\
  David M.~Blei\\
  Columbia University\\
}

\begin{document}

\maketitle

\begin{abstract}
Probabilistic modeling is cyclical: we specify a model, infer its posterior,
and evaluate its performance. Evaluation drives the cycle, as we revise our
model based on how it performs. This requires a metric. Traditionally,
predictive accuracy prevails. Yet, predictive accuracy does not tell the whole
story. We propose to evaluate a model through posterior dispersion. The idea is
to analyze how each datapoint fares in relation to posterior uncertainty around
the hidden structure. We propose a family of \glspl{PDI} that
capture this idea. A \gls{PDI} identifies rich patterns of model mismatch in
three real data examples: voting preferences, supermarket shopping, and
population genetics.
\end{abstract}
 \glsresetall{}

\section{Introduction}
\label{sec:introduction}

Probabilistic modeling is a flexible approach to analyzing structured data.
Three steps define the approach. First, specify a model; this
captures our structural assumptions about the data. Then, we infer the hidden
structure; this means computing (or approximating) the posterior. Last, we
evaluate the model; this helps build better models down the road. 

How do we evaluate models? Decades of reflection have led to
deep and varied forays into model checking, comparison, and criticism
\citep{gelman2013bayesian}. But a common theme permeates all approaches to model
evaluation: the desire to generalize well. 

In machine learning, we traditionally employ two
complementary tools: predictive accuracy and cross-validation.
Predictive accuracy is the target evaluation metric. Cross-validation captures a
notion of generalization and justifies holding out data. This simple 
combination has fueled the development of myriad probabilistic models
\citep{bishop2006pattern,murphy2012machine}. 

Does predictive accuracy tell the whole story? Predictive accuracy
at the datapoint level offers a way to evaluate each observation in the
dataset.
In this way, pointwise predictive accuracy indicates datapoints that do not
match the model well. Yet it does not tell us \emph{how} the mismatch occurs.

\parhead{Main idea.}
We propose to evaluate probabilistic models through the idea of posterior
dispersion, analyzing how each datapoint fares in relation to posterior
uncertainty
around the hidden structure. To capture this, we propose a family of \glspl
{PDI}. These are per-datapoint quantities, each a variance to mean ratio of
the model likelihood with respect to the posterior. A \gls{PDI} highlights
datapoints that exhibit the most uncertainty under the posterior.

Consider a model $p(\mbx,\mbtheta)$ and the likelihood of a datapoint $p(x_n
\mid \mbtheta)$. It depends on some hidden structure $\mbtheta$ that we seek to
uncover. Since $\mbtheta$ is random, we can view the likelihood as a random
variable and ask: how does the likelihood of $x_n$ vary with respect to the
posterior $p(\mbtheta\mid\mbx)$? 

To answer this, we appeal to various forms of dispersion, such as the variance
of the likelihood under the posterior. We propose a family of calibrated
dispersion criteria of form
\begin{align*}
  \textsc{pdi}
  &=
  \frac
  {\text{variance of likelihood under posterior}}
  {\text{mean of likelihood under posterior}}
  =
  \frac
  {\V_{\mbtheta \vert \mbx}[p(x_n\mid\mbtheta)]}
  {\E_{\mbtheta \vert \mbx}[p(x_n\mid\mbtheta)]}.
\end{align*}
Here is a mental picture. Consider a study of human body temperatures. The
posterior represents
uncertainty around temperature. Imagine a high measurement $T_\text{fever}$. 
Its likelihood varies rapidly across plausible values for $T$. 
This datapoint is well modeled, but is sensitive to the posterior. Now imagine a
zero measurement. This datapoint is poorly modeled, but consistently so: the
thermometer is broken. A \gls{PDI} highlights the first datapoint as a
more important type of mismatch than the second.

\begin{center}
\includegraphics[height=0.5in]{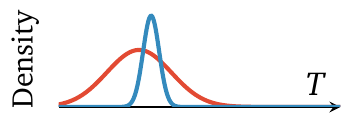}
\includegraphics[height=0.6in]{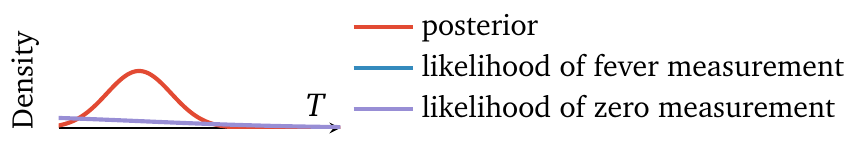}
\end{center}

Section\nobreakspace \ref {sec:experimental} presents
an empirical study of model mismatch in three real-world examples: voting
preferences, supermarket shopping, and population genetics. In each of these
cases, a \gls{PDI} provides insight into the model and offers concrete
directions for improvement.

\parhead{Related work.} This paper relates to a constellation of ideas from
statistical model criticism. \gls{PDI} bears similarity to \textsc
{anova}, which is a frequentist approach to evaluating explanatory variables
in linear regression \citep{davison2003statistical}. 
\citet{gelman1996posterior} cemented the
idea of studying predictive accuracy of probabilistic models at the data level; 
\citet{vehtari2012survey} and \citet{betancourt2015unified} give up-to-date
overviews of these ideas. Recent forays into model criticism, such as 
\citet{gelman2014understanding},
\citet{vehtari2014bayesian} and \citet{piironen2015comparison},
explore the relationship between cross-validation and information criteria, such
as \gls{WAIC} \citep{watanabe2010asymptotic,vehtari2016practical}.
\gls{WAIC} offers an intuitive connection to cross validation
\citep{vehtari2002bayesian,watanabe2015bayesian}; we draw inspiration from it
in this paper too. In the machine learning community, 
\citet{gretton2007kernel} and \citet{chwialkowski2016kernel} developed effective
kernel-based methods for independence and goodness-of-fit tests. Recently, 
\citet{lloyd2015statistical} visualized smooth regions of
data space that the model fails to explain. In contrast, we focus directly on
the datapoints, which can live in high-dimensional spaces that are difficult to
visualize.

 \section{Posterior Dispersion Indices}
\label{sec:pdi}

\glsreset{PDI}
A \gls{PDI} highlights datapoints that exhibit the most
uncertainty with respect to the hidden structure of a model. Here is the
road map for this section. A small case study
illustrates how a \gls{PDI} gives more insight beyond predictive accuracy. 
Definitions, theory, and another small analysis give further insight; a
straightforward
algorithm leads into the empirical study.

\subsection{A motivating case study: 44\% outliers?}
\label{sub:44outlier}

\begin{figure}[!htb]
\centering
\includegraphics[width=5.5in]{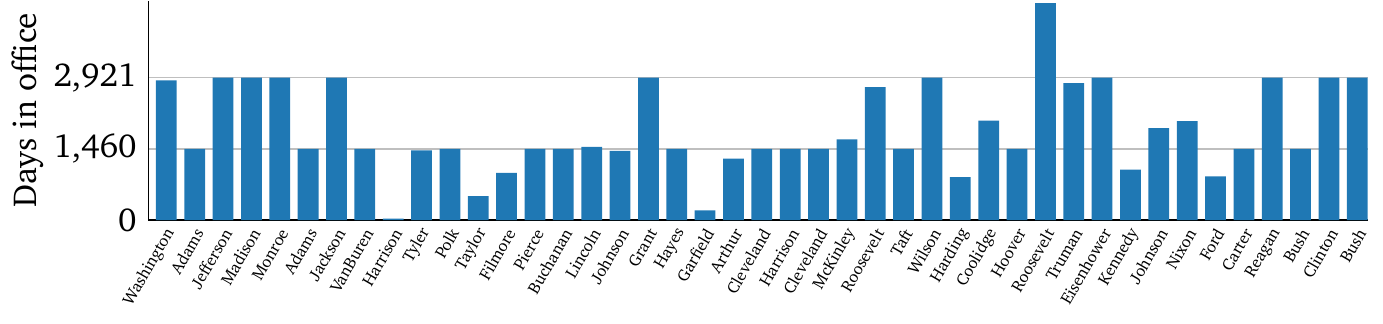}
\vspace*{-16pt}
\caption{The number of days each U.S.~president stayed in office. Typical
durations are easy to identify; gray lines indicate one- and two-term stays.
Appendix\nobreakspace \ref {app:us_presidents} presents numerical values.}
\label{fig:presidents_data}
\end{figure}

\citet{hayden2005dataset} considers the number of days each U.S.~president stayed
in office. Figure\nobreakspace \ref {fig:presidents_data} plots the data. One-term presidents stay
in office for around 1460 days; two-term presidents approximately double that.
Yet many presidents deviate from this ``two bump'' trend.\footnote{ 
\citet{hayden2005dataset} submits that 44\% of presidents may be outliers.}

A reasonable model for such data is a mixture of negative binomial
distributions.\footnote{
A Poisson likelihood is too underdispersed.
}
Consider the ``second'' parameterization of the negative
binomial with mean $\mu$ and variance $\mu+\nicefrac{\mu^2}{\phi}$.
Posit gamma priors on the (non-negative) latent variables.
Set the prior on $\mu$ to match the mean and variance of the data
\citep{robbins1964empirical}. Choose an uninformative prior on $\phi$. Three
mixtures make sense: two for the typical trends and one for the rest. 

The complete model is
\begin{align*}
  p(\mbpi) &= \text{Dirichlet}(\mbpi\prm\alpha=(1,1,1))\\
  p(\mbmu) &= \prod_{k=1}^3
  \Gam(\mu_k\prm\text{mean and variance matched to that of data})\\
  p(\mbphi) &= \prod_{k=1}^3
  \Gam(\phi_k\prm a=1, \beta=0.01)\\
  p(x_n\mid\mbpi,\mbmu,\mbphi) 
  &= 
  \sum_{k=1}^3
  \pi_k
  \text{NB2}(x_n\prm\mu_k,\phi_k).
\end{align*}

Fitting this model gives posterior mean estimates 
$\smash{\widehat{\mbmu} = (1461, 2896, \red{1578})}$ with corresponding 
$\smash{\widehat{\mbphi} = (470,  509,  \red{1.3})}$. The first two clusters
describe the
two typical term durations, while the third (highlighted in \red{red}) is a
dispersed negative binomial that attempts to describe the rest of the data.

\begin{figure}[!htb]
\centering
\includegraphics[width=5.5in]{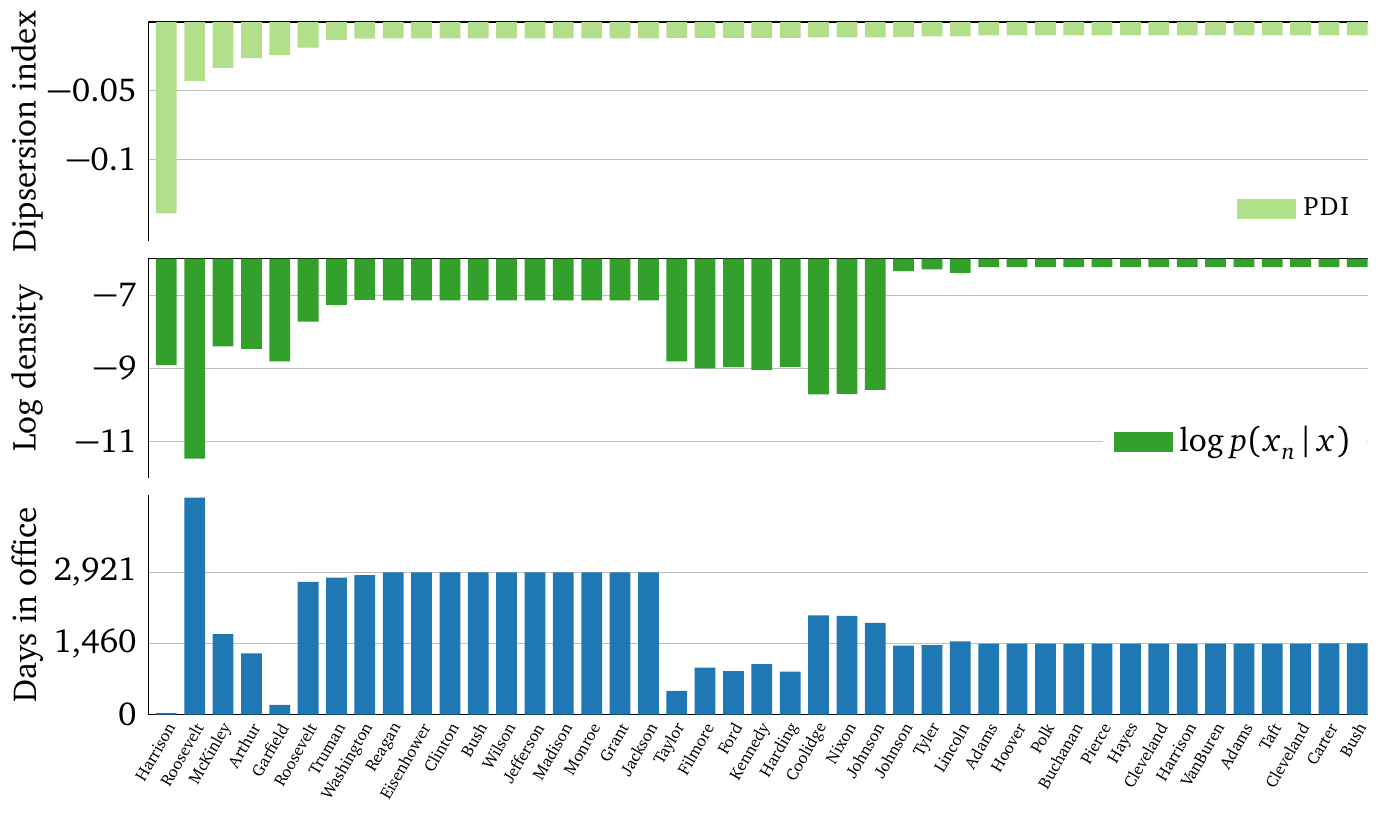}
\vspace*{-16pt}
\caption{\gls{PDI} and log predictive accuracy of each president under a
mixture of three negative binomials model. Presidents sorted by \gls{PDI}. 
The closer to zero, the better. (Code in supplement.)
}
\label{fig:presidents_sorted}
\end{figure}

We compute a \gls{PDI} (defined in Section\nobreakspace \ref {sub:definitions})
and the posterior predictive density for each president
$p(x_n\mid\mbx)$. Figure\nobreakspace \ref {fig:presidents_sorted} compares both metrics and sorts
the presidents according to the \gls{PDI}.

Some presidents are clear outliers: 
Harrison \texttt{\color{Gray}[31:natural death]},
Roosevelt \texttt{\color{Gray}[4452:three terms]}, 
and Garfield \texttt{\color{Gray}[199:assassinated]}. 
However, there are 3 presidents with worse predictive accuracy than Harrison:
Coolidge, Nixon, and Johnson. A \gls{PDI} differentiates Harrison from these
three by also taking into account the variance of his likelihood with respect to
the posterior.

\gls{PDI} also calls attention to 
McKinley \texttt{\color{Gray}[1655:assassinated]} and
Arthur \texttt{\color{Gray}[1260:succeeded Garfield]}; this is because 
they are close to the peaked negative binomial cluster at 1460 but not close
enough to have good predictive accuracy. They are datapoints whose
likelihoods are rapidly changing with respect to the posterior, like the fever 
measurement in the introduction.

This case study suggests that predictive probability does not tell the entire
story. Datapoints can exhibit low predictive accuracy in different ways. 
The following two sections define and explain how a \gls{PDI} captures this
effect.

\subsection{Definitions}
\label{sub:definitions}

Let $\mbx = \{x_n\}_1^N$ be a dataset with $N$ observations. A 
probabilistic model
has two parts. The first is the likelihood, $p(x_n \mid \mbtheta)$. It relates
an observation $x_n$ to hidden patterns described by a set latent random
variables $\mbtheta$. If the observations are independent and identically
distributed, the likelihood of the dataset factorizes as $p(\mbx \mid \mbtheta)
= \prod_{n} p(x_n \mid \mbtheta)$.

The second is the prior density, $p(\mbtheta)$. It captures the structure we
expect from the hidden patterns. Combining the likelihood and the
prior gives the joint density $p(\mbx, \mbtheta) = p(\mbx \mid \mbtheta) 
p (\mbtheta)$. Conditioning the joint on the observed data gives the posterior  
density, $p(\mbtheta \mid \mbx)$.

Treat the likelihood of each datapoint as a function of $\mbtheta$.
To evaluate the model, we analyze how each datapoint fares in
relation to the posterior density. Consider these expectations and variances
with respect to the posterior,
\begin{align}
\begin{aligned}
  \mu(n) &= \E_{\mbtheta \vert \mbx}[p(x_n\mid\mbtheta)]
  &\;
  \mu_\text{log}(n) &= \E_{\mbtheta \vert \mbx}[\log p(x_n\mid\mbtheta)],\\
  \sigma^2(n) &= \V_{\mbtheta \vert \mbx}[p(x_n\mid\mbtheta)]
  &\;
  \sigma_\text{log}^2(n) &= \V_{\mbtheta \vert \mbx}[\log p(x_n\mid\mbtheta)].
\end{aligned}
\label{eq:mean_var_def}
\end{align}

Each includes the likelihood in a slightly different fashion. The first
expectation is a familiar object: $\mu(n)$ is the posterior predictive
distribution. 

A \gls{PDI} is a ratio of these variances to expectations. Taking the ratio
calibrates this quantity for each datapoint. Recall the mental picture from the
introduction. The variance of the likelihood under the posterior highlights
potential model mismatch; dividing by the mean calibrates this spread to its
predictive accuracy. 

Related ratios also appear
in classical statistics under a variety of forms, such as indices of dispersion
\citep{hoel1943indices}, coefficients of variation 
\citep{koopmans1964confidence}, or the Fano factor
\citep{fano1947ionization}. They all quantify dispersion of samples from
a random process.

\glsreset{WAPDI}
In this paper, we propose and study the \gls{WAPDI}, 
\begin{align*}
  \textsc{wapdi}(n)
  &=
  \frac
  {\sigma_\text{log}^2(n)}
  {\log \mu(n)}.
\end{align*}

Its form and name comes from the widely applicable information criterion
\begin{align*}
\textsc{waic}
&=
-\frac{1}{N}\sum_n \log \mu(n) + \sigma_\text{log}^2(n).
\end{align*}
\gls{WAIC} measures generalization error; it asymptotically
equates to leave-one-one cross validation \citep{watanabe2010asymptotic,watanabe2015bayesian}.
\gls{WAPDI} has two advantages; both are practically
motivated. First, we hope the reader is computing some estimate of
generalization error. \citet{gelman2014understanding} recommends
\gls{WAIC}, since it is easy to compute and
designed for common machine learning models 
\citep{watanabe2010asymptotic}. Computing \gls{WAIC} gives \gls{WAPDI}
for free. Second, the variance is a second-order moment
calculation; using the log likelihood gives numerical stability to the 
computation. (More on computation in Section\nobreakspace \ref {sub:computation}.)

\gls{WAPDI} compares the variance of the log likelihood to the log posterior
predictive. This gives insight into \emph{how} the likelihood of a datapoint
fares under the posterior distribution of the hidden patterns. We now analyze
this in more detail.

\subsection{Intuition: not all predictive probabilities are created equal}
\label{sub:not_all}

The posterior predictive density is an expectation, 
$\E_{\mbtheta \vert \mbx}[p(x_\text{new}\mid\mbtheta)]
= 
\int\!\!
p(x_\text{new}\mid\mbtheta) p(\mbtheta\mid \mbx) \dif\mbtheta.
$
Expectations are integrals: areas under a curve. Different likelihood and
posterior combinations can lead to similar integrals.

A toy model illustrates this. Consider a gamma likelihood with fixed shape,
and place a gamma prior on the rate. The model is
\begin{align*}
p(\beta) &= \Gam(\beta\prm a_0=1, b_0=1),\\
p(\mbx\mid\beta) &= \prod_{n=1}^N\Gam(x_n\prm a=5, \beta),
\end{align*}
which gives the posterior
$
p(\beta\mid \mbx) 
= 
\Gam(\beta\prm a=a_0+2N, b = b_0+\sum_n x_n)
$.

\begin{figure}[!htb]
\centering
\includegraphics[width=5.5in]{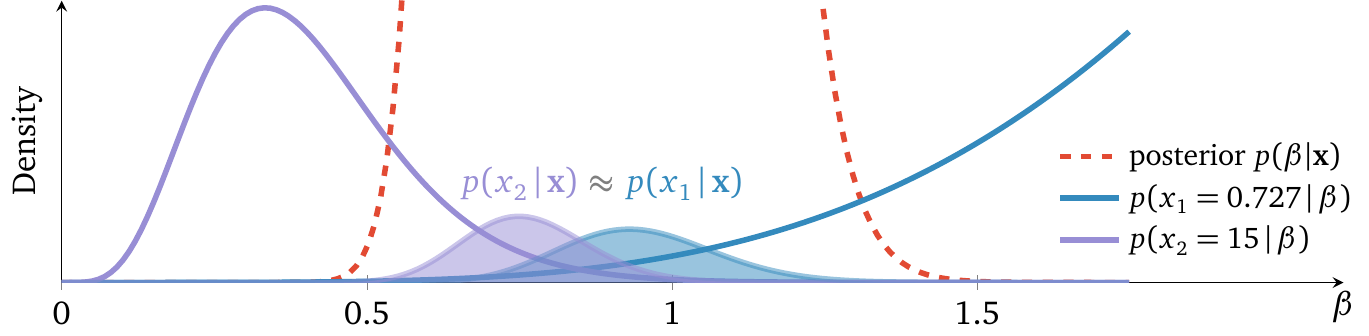}
\vspace*{-6pt}
\caption{Not all predictive probabilities are created equal. The translucent
curves are the two likelihoods multiplied by the posterior (cropped).
The posterior predictive $p(x_{1,2}\mid\mbx)$ for each datapoint is the area
under
the curve. While
both datapoints have the same predictive accuracy, the likelihood for $x_2$
has higher variance under the posterior; $x_2$ is more sensitive to the spread 
of the posterior than $x_1$. \gls{WAPDI} captures this effect. (Code in
supplement.)}
\label{fig:not_all_same}
\end{figure}

Now simulate a dataset of size $N=10$ with $\beta=1$; the data have mean 
$\nicefrac{a}{\beta} = 5$. Now consider an outlier at $15$.
We can find another $x$ value with essentially the same
predictive accuracy
\begin{align*}
  \log p(x_1=0.727\mid \mbx)
  &=
  -5.633433,
  &
  \log p(x_2=15\mid \mbx)
  &=
  -5.633428.
\intertext{Yet their \gls{WAPDI} values differ by an order of magnitude}
  \textsc{wapdi}(x_1=0.727)
  &=
  -0.067,
  &
  \textsc{wapdi}(x_2=15)
  &=
  -0.229.
\end{align*}
In this case, \gls{WAPDI} highlights $x_2=15$ as a more severe outlier than
$x_1=0.727$, even though they have the same predictive accuracy. What does that
mean? Figure\nobreakspace \ref {fig:not_all_same} depicts the difference.

The following lemma explains how \gls{WAPDI} measures this effect.

\begin{lemma}
If $\log p(x_n\mid\mbtheta)$ is at least twice differentiable and the
posterior $p(\mbtheta\mid\mbx)$ has finite first and second moments, then a
first-order Taylor expansion gives
\begin{align}
  \textsc{wapdi}(n)
  &\approx
  \frac
  {\left(\log p^\prime(x_n\mid \E_{\mbtheta \vert \mbx}[\mbtheta])\right)^2
  \V_{\mbtheta \vert \mbx}[\mbtheta]}
  {\log \E_{\mbtheta \vert \mbx}[p(x_n\mid\mbtheta)]}.
\label{eq:wapdi_taylor}
\end{align}
\end{lemma}

\begin{corollary}
\gls{WAPDI} highlights datapoints whose likelihood is rapidly changing at the
posterior mean estimate of the latent variables. 
($\V_{\mbtheta \vert \mbx}[\mbtheta]$ is constant across $n$.)
\end{corollary}

Looking back at Figure\nobreakspace \ref {fig:not_all_same}, the likelihood $p(x_2=15\mid\beta)$ 
indeed changes rapidly under the posterior. \gls{WAPDI} reports the ratio of
this rate-of-change to the area under the curve. In this special example, only
the numerator matters, since the denominator is the same for both
datapoints.

\begin{corollary}
Equation\nobreakspace \textup {(\ref {eq:wapdi_taylor})} is zero if and only if the
posterior mean coincides with the maximum likelihood estimate of $\mbtheta$.
($\V_{\mbtheta \vert \mbx}[\mbtheta]$ is positive for finite $N$.)
\end{corollary}

For most interesting models, we do not expect such a coincidence. However, in
practice, we find \gls{WAPDI} to be close to zero for datapoints that match
the model well.

\subsection{Computation}
\label{sub:computation}

Calculating \gls{WAPDI} is straightforward. The only requirement are samples
from the posterior. This is precisely the output of an \gls{MCMC} sampling
algorithm. (We used the no-U-turn sampler \citep{hoffman2014nuts} for the 
analyses above.) Other inference procedures, such as variational inference, give
an analytic approximation to the posterior \citep{jordan1999introduction,
blei2016variational}. Drawing
samples from an approximate posterior also works.

Equipped with $S$ samples from the posterior, Monte Carlo
integration \citep{robert1999monte} gives unbiased estimates of the quantities 
in Equation\nobreakspace \textup {(\ref {eq:mean_var_def})}. The variance of these estimates decreases as
$\mathcal{O}(\nicefrac{1}{S})$; we assume $S$ is sufficiently large to cover the
posterior \citep{gelman2014understanding}. We default to $S=1000$ in our
experiments. Algorithm\nobreakspace \ref {alg:wapdi} summarizes these steps.

\begin{center}
\parbox{4in}{
\begin{algorithm}[H]
  \caption{Calculating posterior dispersion indices.}
  \SetAlgoLined
  \DontPrintSemicolon
  \BlankLine
  \KwIn{Data $\mbx = \{\mbx_n\}_1^N$, model $p(\mbx,\mbtheta)$.}
  \BlankLine
  \KwOut{Posterior dispersion index for each datapoint $x_n$.}
  \BlankLine
  Draw $S$ samples $\{\mbtheta\}_1^S$ from posterior (approximation) $p
  (\mbtheta\mid\mbx)$.
  \BlankLine
  \For{n \textnormal{ in } 1, 2, \ldots, N}{
    \BlankLine
    Estimate $\mu(n), \mu_\text{log}(n), 
    \sigma^2(n), \sigma_\text{log}^2(n)$ from samples $\{\mbtheta\}_1^S$.
    \BlankLine
    Store these estimates.
    \BlankLine
    Return the desired ratio.
    \BlankLine
  }
\label{alg:wapdi}
\end{algorithm}
}
\end{center}

 \section{Experimental Study}
\label{sec:experimental}

We now explore three real data examples using modern machine learning models: 
voting preferences, supermarket shopping, and population genetics.

\subsection{Voting preferences: a hierarchical logistic regression model}
In 1988, CBS conducted a U.S.~nation-wide survey of voting preferences. People
indicated their preference towards the Democratic or Republican 
presidential candidate. Each individual also declared their gender, age,
race, education level, and the state they live in; $11\,566$ individuals
participated.

\citet{gelman2006data} study this data through a hierarchical logistic
regression model. They begin by modeling gender, race, and state; the state
variable has a hierarchical prior. This model is easy to fit using 
\gls{ADVI} within Stan \citep{kucukelbir2015automatic,carpenter2015stan}. 
(Model and inference details in Appendix\nobreakspace \ref {app:hier_log_reg}.)

\begin{table}[!htbp]
\setlength{\tabcolsep}{2pt}
\centering 
\sc
\parbox{.49\linewidth}{
\centering
\begin{tabular}{rcccccccccc}
\toprule
\textbf{vote}   & r  & r  & r  & r  & r  & r  & r  & r  & r  & r\\
\textbf{sex} & f  & f  & f  & f  & f  & f  & f  & f  & f  & f\\
\textbf{race}   & b  & b  & b  & b  & b  & b  & b  & b  & b  & b\\
\textbf{state}  & {wa} & {wa} & {ny} & {wi} & {ny} & {ny} & {ny} & {ny} & {ma} & {ma}\\
\bottomrule
\end{tabular}
\caption{Lowest predictive accuracy}
\label{tab:lowest_logpred}}
\parbox{.49\linewidth}{ 
\centering
\begin{tabular}{rcccccccccc}
\toprule
\textbf{vote}   & d  & d  & d  & d  & d  & d  & r  & d  & d  & d\\
\textbf{sex} & f  & f  & f  & f  & f  & m  & m  & m  & m  & m\\
\textbf{race}   & w  & w  & w  & w  & w  & w  & w  & w  & w  & w\\
\textbf{state}  & {wy} & {wy} & {wy} & {wy} & {wy} & {wy} & {wy} & {dc} & {dc} &
{nv}\\
\bottomrule
\end{tabular}
\caption{Lowest \gls{WAPDI}}
\label{tab:lowest_WAPDI}}

\end{table}
 
Tables\nobreakspace \ref {tab:lowest_logpred} and\nobreakspace  \ref {tab:lowest_WAPDI} show the individuals with the lowest 
predictive accuracy and \gls{WAPDI}. The nation-wide trend
predicts that females (\textsc{f}) who identify as black (\textsc{b}) have a
strong preference to vote democratic (\textsc{d}); predictive accuracy
identifies the few individuals who defy this trend. However, there is not much
to do with this information; the model identifies a nation-wide trend that
correctly describes most female black voters. In contrast, \gls{WAPDI} points to
parts of the dataset that the model fails to describe; these are datapoints
that we might try to explain better with a revised model.

Most of the individuals with low \gls{WAPDI} live in
Wyoming, the District of Columbia, and Nevada. We focus on Wyoming
and Nevada. The average \gls{WAPDI} for Wyoming and Nevada are $-0.057$ and
$-0.041$; these are baselines that we seek to improve. (The closer to zero, the
better.)

Consider expanding the model by modeling age. Introducing age into the model
with a hierarchical prior reveals that older voters tend to vote Republican.
This helps explain Wyoming voters; their average \gls{WAPDI} drops down from
$-0.057$ to $-0.04$; however Nevada's average \gls{WAPDI} remains unchanged. 
This means that Nevada's voters may not follow the national
age-dependent trend. Now consider removing age and introducing education in a
similar way. Education helps explain voters from both states; the average \gls
{WAPDI} for Wyoming and Nevada drop to $-0.041$ and $-0.029$.

\gls{WAPDI} thus captures interesting datapoints beyond what predictive accuracy
reports. As expected, predictive accuracy still highlights the same female black
voters in both expanded models; \gls{WAPDI} illustrates a deeper way to evaluate
this model.

\subsection{Supermarket shopping: a hierarchical Poisson factorization model}
Market research firm IRi hosts an anonymized dataset of customer shopping 
behavior at U.S.~supermarkets \citep{bronnenberg2008database}. The 
dataset tracks $136\,584$ ``checkout'' sessions; each session contains a 
basket of purchased items. An inventory of $7\,903$ items range across
categories such as carbonated beverages, toiletries, and yogurt. 

What items do customers tend to purchase together? To study this, consider
a \gls{HPF} model \citep{gopalan2015scalable}. \gls{HPF} models the quantities
of items purchased in each session with a Poisson likelihood; its rate is 
an inner product between a session's preferences $\mbtheta_s$ and the item
attributes $\mbbeta$. Hierarchical priors on $\mbtheta$ and $\mbbeta$
simultaneously promote sparsity, while accounting for variation in
session size and item popularity. Some sessions contain only a few items;
others are large purchases. (Model and inference details in 
Appendix\nobreakspace \ref {app:hpf}.)

A $20$-dimensional \gls{HPF} model discovers intuitive trends. A few stand out.
Snack-craving customers like to buy Doritos tortilla chips along with Lay's
potato chips. Morning birds typically pair Cheerios cereal with 2\% skim milk.
Yoplait fans tend to purchase many different flavors at the same time. 
Tables\nobreakspace \ref {tab:iri_morningbird} and\nobreakspace  \ref {tab:iri_yoplait} show the top five items in two of
these twenty trends.

\begin{table}[!htbp]
\centering 
\small
\parbox{.49\linewidth}{
\centering
\begin{tabular}{ll}
\toprule
\textbf{Item Description} & \textbf{Category}\\
\midrule
Brand A: 2\% skim milk             & milk rfg skim/lowfat  \\
Cheerios: cereal                   & cold cereal \\
Diet Coke: soda                    & carbonated beverages \\
Brand B: 2\% skim milk             & milk rfg skim/lowfat \\
Brand C: 2\% skim milk        & milk rfg skim/lowfat \\
\bottomrule
\end{tabular}
\caption{Morning-bird trend}
\label{tab:iri_morningbird}}
\parbox{.49\linewidth}{ 
\centering
\begin{tabular}{ll}
\toprule
\textbf{Item Description} & \textbf{Category}\\
\midrule
Yoplait: raspberry flavor    & yogurt rfg \\
Yoplait: peach flavor    & yogurt rfg \\
Yoplait: strawberry flavor  & yogurt rfg \\
Yoplait: blueberry flavor    & yogurt rfg \\
Yoplait: blackberry flavor    & yogurt rfg \\
\bottomrule
\end{tabular}
\caption{Yoplait fan trend}
\label{tab:iri_yoplait}}

\end{table}

 Sessions where a customer purchases many items from different
categories have low predictive accuracy. This makes sense as these
customers do not exhibit a trend; mathematically, there is no combination
of item attributes $\mbbeta$ that explain buying items from disparate
categories. For example, the session with the lowest predictive accuracy
contains 117 items ranging from coffee to hot dogs.

\gls{WAPDI} highlights an entirely different aspect of the \gls{HPF} model.
Sessions with low \gls{WAPDI} contain similar items but exhibit many
purchases of a single item. Table\nobreakspace \ref {tab:lowest_WAPDI_session} shows an example of
a session where a customer purchased 14 blackberry Yoplait yogurts, but only a
few of the other flavors.

\begin{table}[!htbp]
\centering 
\small
\begin{tabular}{lc}
\toprule
\textbf{Item Description} & \textbf{Quantity}\\
\midrule
Yoplait: blackberry flavor & 14 \\
Yoplait: strawberry flavor  & 2 \\
Yoplait: raspberry flavor  & 2 \\
Yoplait: peach flavor  & 1 \\
Yoplait: cherry flavor  & 1 \\
Yoplait: mango flavor  & 1 \\
\bottomrule
\end{tabular}
\caption{A session with low \gls{WAPDI}}
\label{tab:lowest_WAPDI_session}
\end{table}

This indicates that the Poisson likelihood assumption may not be flexible enough
to model customer purchasing behavior. Perhaps a negative binomial likelihood
could model this kind of spiked activity better. Another option might be to
keep the Poisson likelihood but increase the hierarchy of the probabilistic
model; this approach may identify item attributes that
explain such purchases. In either case, \gls{WAPDI} identifies a
valuable aspect of the data that the \gls{HPF} struggles to capture: sessions
with spiked activity. This is a concrete direction for model revision.

\subsection{Population genetics: a mixed membership model}

Do all people who live nearby have similar genomes? Not necessarily. 
Population genetics considers how individuals exhibit
ancestral patterns of mutations. Begin with $N$ individuals and 
$L$ locations on the genome. For each location, report whether each individual
reveals a mutation. This gives an ($N \times L$) dataset $\mbx$ where $x_{nl}
\in {0,1,2,3}$. (We assume two specific forms of mutation; 3 encodes a missing
observation.)

Mixed membership models offer a way to study this 
\citep{pritchard2000inference}.
Assume $K$ ancestral populations $\mbphi$; these are the mutation
probabilities of each location. Each individual mixes these populations with
weights $\mbtheta$; these are the mixing proportions. Place a beta prior
on the mutation probabilities and a Dirichlet prior on the mixing proportions. 

\begin{figure}[!htb]
\centering
\includegraphics[width=5.5in]{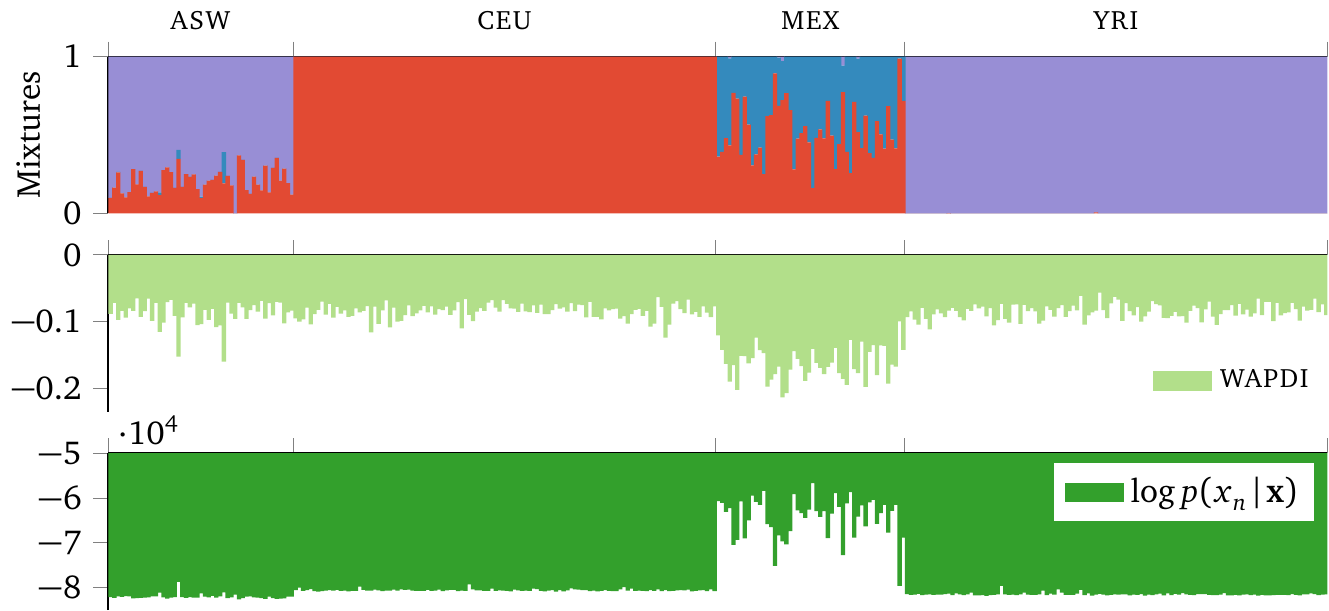}
\caption{Individuals of African ancestry in southwest U.S.~(\textsc{asw}) and
Mexican ancestry in Los Angeles (\textsc{mex}) exhibit a mixture of two
populations. In contrast, Utah residents with European ancestry 
(\textsc{ceu}) and members of the Yoruba group in Nigeria (\textsc{yri}) are
mostly uniform.}
\label{fig:admixture}
\end{figure}

We study a dataset of $N=324$ individuals from four geographic locations 
and focus on $L=13\,928$ locations on the genome.
Figure\nobreakspace \ref {fig:admixture} shows how these
individuals mix $K=3$ ancestral populations.
(Data, model, and inference details in Appendix\nobreakspace \ref {app:admixture}.)

\gls{WAPDI} reveals three interesting patterns of mismatch here. 
First, individuals with low \gls{WAPDI} have many missing observations; 
the bottom 10\% of \gls{WAPDI} have $1\,344$ missing values, in contrast
to $563$ for the lowest 10\% of predictive scores. We may consider
directly modeling these missing observations.

Second, \textsc{asw} has two individuals with low \gls{WAPDI};
their mutation patterns are outliers within the group. While the average
individual reveals 272 mutations away from the median genome, these individuals
show 410 and 383 mutations. This points to potential mishaps while gathering or
pre-processing the data.

Last, \textsc{mex} exhibits good predictive accuracy, yet low \gls{WAPDI}. Based
on predictive accuracy, we may happily accept these patterns. Yet \gls{WAPDI}
highlights a serious issue with the inferred populations. The blue and red 
populations are almost twice as correlated ($0.58$) as the other possible
combinations ($0.24$ and $0.2$). In other words, the blue and red populations
represent similar patterns of mutations at the same locations. These
populations, as they stand, are not necessarily interpretable. Revising the
model to penalize correlation may be a direction worth pursuing.
 
\section{Discussion}
\label{sec:discussion}

\glsreset{PDI}
\glsreset{WAPDI}
A \gls{PDI} identifies informative forms of model mismatch beyond 
predictive accuracy. By highlighting which datapoints exhibit
the most uncertainty under the posterior, a \gls{PDI} offers a new perspective
into evaluating probabilistic models. Here, we show how the \gls{WAPDI} reveals
concrete directions for model improvement across a range of models and
applications.

There are several exciting research directions going forward. One is to extend
the notion of a \gls{PDI} to non-exchangeable data. Another is to leverage the
bootstrap to
extend this idea beyond probabilistic models. Finally, theoretical connections
to cross-validation await discovery, while ideas from importance sampling could
reduce the variance of \gls{PDI} computations.
 
\textbf{Acknowledgments}

We thank David Mimno, Maja Rudolph, Dustin Tran, and Aki Vehtari 
for their insightful comments. This work is supported by NSF IIS-1247664, ONR
N00014-11-1-0651, and DARPA FA8750-14-2-0009.
 \clearpage
\bibliographystyle{apa}
\bibliography{BIB}

\begin{thebibliography}{}

\bibitem[\protect\astroncite{Betancourt}{2015}]{betancourt2015unified}
Betancourt, M. (2015).
\newblock A unified treatment of predictive model comparison.
\newblock {\em arXiv preprint arXiv:1506.02273}.

\bibitem[\protect\astroncite{Bishop}{2006}]{bishop2006pattern}
Bishop, C.~M. (2006).
\newblock {\em Pattern Recognition and Machine Learning}.
\newblock Springer New York.

\bibitem[\protect\astroncite{Blei et~al.}{2016}]{blei2016variational}
Blei, D.~M., Kucukelbir, A., and McAuliffe, J.~D. (2016).
\newblock Variational inference: a review for statisticians.
\newblock {\em arXiv preprint arXiv:1601.00670}.

\bibitem[\protect\astroncite{Bronnenberg
  et~al.}{2008}]{bronnenberg2008database}
Bronnenberg, B.~J., Kruger, M.~W., and Mela, C.~F. (2008).
\newblock The {IRi} marketing data set.
\newblock {\em Marketing Science}, 27(4).

\bibitem[\protect\astroncite{Carpenter et~al.}{2015}]{carpenter2015stan}
Carpenter, B., Gelman, A., Hoffman, M., Lee, D., Goodrich, B., Betancourt, M.,
  Brubaker, M.~A., Guo, J., Li, P., and Riddell, A. (2015).
\newblock Stan: a probabilistic programming language.
\newblock {\em JSS}.

\bibitem[\protect\astroncite{Chwialkowski
  et~al.}{2016}]{chwialkowski2016kernel}
Chwialkowski, K., Strathmann, H., and Gretton, A. (2016).
\newblock A kernel test of goodness of fit.
\newblock {\em arXiv preprint arXiv:1602.02964}.

\bibitem[\protect\astroncite{Davison}{2003}]{davison2003statistical}
Davison, A.~C. (2003).
\newblock {\em Statistical models}.
\newblock Cambridge University Press.

\bibitem[\protect\astroncite{Fano}{1947}]{fano1947ionization}
Fano, U. (1947).
\newblock Ionization yield of radiations {II}.
\newblock {\em Physical Review}, 72(1):26.

\bibitem[\protect\astroncite{Gelman et~al.}{2013}]{gelman2013bayesian}
Gelman, A., Carlin, J.~B., Stern, H.~S., Dunson, D.~B., Vehtari, A., and Rubin,
  D.~B. (2013).
\newblock {\em {B}ayesian Data Analysis}.
\newblock CRC Press.

\bibitem[\protect\astroncite{Gelman and Hill}{2006}]{gelman2006data}
Gelman, A. and Hill, J. (2006).
\newblock {\em Data analysis using regression and multilevel/hierarchical
  models}.
\newblock Cambridge University Press.

\bibitem[\protect\astroncite{Gelman et~al.}{2014}]{gelman2014understanding}
Gelman, A., Hwang, J., and Vehtari, A. (2014).
\newblock Understanding predictive information criteria for {B}ayesian models.
\newblock {\em Statistics and Computing}, 24(6):997--1016.

\bibitem[\protect\astroncite{Gelman et~al.}{1996}]{gelman1996posterior}
Gelman, A., Meng, X.-L., and Stern, H. (1996).
\newblock Posterior predictive assessment of model fitness via realized
  discrepancies.
\newblock {\em Statistica sinica}, 6(4):733--760.

\bibitem[\protect\astroncite{Gopalan et~al.}{2015}]{gopalan2015scalable}
Gopalan, P., Hofman, J.~M., and Blei, D.~M. (2015).
\newblock Scalable recommendation with hierarchical {P}oisson factorization.
\newblock {\em UAI}.

\bibitem[\protect\astroncite{Gretton et~al.}{2007}]{gretton2007kernel}
Gretton, A., Fukumizu, K., Teo, C.~H., Song, L., Sch{\"o}lkopf, B., and Smola,
  A.~J. (2007).
\newblock A kernel statistical test of independence.
\newblock In {\em NIPS}.

\bibitem[\protect\astroncite{Hayden}{2005}]{hayden2005dataset}
Hayden, R.~W. (2005).
\newblock A dataset that is 44\% outliers.
\newblock {\em J Stat Educ}, 13(1).

\bibitem[\protect\astroncite{Hoel}{1943}]{hoel1943indices}
Hoel, P.~G. (1943).
\newblock On indices of dispersion.
\newblock {\em The Annals of Mathematical Statistics}, 14(2):155--162.

\bibitem[\protect\astroncite{Hoffman and Gelman}{2014}]{hoffman2014nuts}
Hoffman, M.~D. and Gelman, A. (2014).
\newblock The {N}o-{U}-{T}urn sampler.
\newblock {\em JMLR}, 15(1):1593--1623.

\bibitem[\protect\astroncite{Jordan et~al.}{1999}]{jordan1999introduction}
Jordan, M.~I., Ghahramani, Z., Jaakkola, T.~S., and Saul, L.~K. (1999).
\newblock An introduction to variational methods for graphical models.
\newblock {\em Machine Learning}, 37(2):183--233.

\bibitem[\protect\astroncite{Koopmans et~al.}{1964}]{koopmans1964confidence}
Koopmans, L.~H., Owen, D.~B., and Rosenblatt, J. (1964).
\newblock Confidence intervals for the coefficient of variation for the normal
  and log normal distributions.
\newblock {\em Biometrika}, 51(1/2):25--32.

\bibitem[\protect\astroncite{Kucukelbir et~al.}{2015}]{kucukelbir2015automatic}
Kucukelbir, A., Ranganath, R., Gelman, A., and Blei, D.~M. (2015).
\newblock Automatic variational inference in {S}tan.
\newblock {\em NIPS}.

\bibitem[\protect\astroncite{Lloyd and Ghahramani}{2015}]{lloyd2015statistical}
Lloyd, J.~R. and Ghahramani, Z. (2015).
\newblock Statistical model criticism using kernel two sample tests.
\newblock In {\em NIPS}.

\bibitem[\protect\astroncite{Mimno et~al.}{2015}]{mimno2015posterior}
Mimno, D., Blei, D.~M., and Engelhardt, B.~E. (2015).
\newblock Posterior predictive checks to quantify lack-of-fit in admixture
  models of latent population structure.
\newblock {\em Proceedings of the National Academy of Sciences}, 112(26).

\bibitem[\protect\astroncite{Murphy}{2012}]{murphy2012machine}
Murphy, K.~P. (2012).
\newblock {\em Machine Learning: a Probabilistic Perspective}.
\newblock MIT Press.

\bibitem[\protect\astroncite{Piironen and
  Vehtari}{2015}]{piironen2015comparison}
Piironen, J. and Vehtari, A. (2015).
\newblock Comparison of {B}ayesian predictive methods for model selection.
\newblock {\em arXiv preprint arXiv:1503.08650}.

\bibitem[\protect\astroncite{Pritchard et~al.}{2000}]{pritchard2000inference}
Pritchard, J.~K., Stephens, M., and Donnelly, P. (2000).
\newblock Inference of population structure using multilocus genotype data.
\newblock {\em Genetics}, 155(2):945--959.

\bibitem[\protect\astroncite{Raj et~al.}{2014}]{raj2014faststructure}
Raj, A., Stephens, M., and Pritchard, J.~K. (2014).
\newblock {fastSTRUCTURE}: variational inference of population structure in
  large {SNP} data sets.
\newblock {\em Genetics}, 197(2):573--589.

\bibitem[\protect\astroncite{Robbins}{1964}]{robbins1964empirical}
Robbins, H. (1964).
\newblock The empirical {B}ayes approach to statistical decision problems.
\newblock {\em Annals of Mathematical Statistics}.

\bibitem[\protect\astroncite{Robert and Casella}{1999}]{robert1999monte}
Robert, C.~P. and Casella, G. (1999).
\newblock {\em Monte Carlo statistical methods}.
\newblock Springer.

\bibitem[\protect\astroncite{Vehtari et~al.}{2016}]{vehtari2016practical}
Vehtari, A., Gelman, A., and Gabry, J. (2016).
\newblock Practical {B}ayesian model evaluation using leave-one-out
  cross-validation and {WAIC}.
\newblock {\em arXiv preprint arXiv:1507.04544}.

\bibitem[\protect\astroncite{Vehtari and Lampinen}{2002}]{vehtari2002bayesian}
Vehtari, A. and Lampinen, J. (2002).
\newblock {B}ayesian model assessment and comparison using cross-validation
  predictive densities.
\newblock {\em Neural Computation}, 14(10):2439--2468.

\bibitem[\protect\astroncite{Vehtari et~al.}{2012}]{vehtari2012survey}
Vehtari, A., Ojanen, J., et~al. (2012).
\newblock A survey of {B}ayesian predictive methods for model assessment,
  selection and comparison.
\newblock {\em Statistics Surveys}, 6:142--228.

\bibitem[\protect\astroncite{Vehtari et~al.}{2014}]{vehtari2014bayesian}
Vehtari, A., Tolvanen, V., Mononen, T., and Winther, O. (2014).
\newblock Bayesian leave-one-out cross-validation approximations for gaussian
  latent variable models.
\newblock {\em arXiv preprint arXiv:1412.7461}.

\bibitem[\protect\astroncite{Watanabe}{2010}]{watanabe2010asymptotic}
Watanabe, S. (2010).
\newblock Asymptotic equivalence of {B}ayes cross validation and widely
  applicable information criterion in singular learning theory.
\newblock {\em JMLR}, 11:3571--3594.

\bibitem[\protect\astroncite{Watanabe}{2015}]{watanabe2015bayesian}
Watanabe, S. (2015).
\newblock {B}ayesian cross validation and {WAIC} for predictive prior design in
  regular asymptotic theory.
\newblock {\em arXiv preprint arXiv:1503.07970}.

\end{thebibliography}

\clearpage
\appendix
\section{U.S.~presidents dataset}
\label{app:us_presidents}

\citet{hayden2005dataset} studies this dataset (and provides a fascinating
educational perspective too).
\begin{table}[!htbp]
\centering  
\begin{tabular}{rl}
\toprule
\textbf{President}  & \textbf{Days}\\
\midrule
Washington & 2864\\
Adams      & 1460\\
Jefferson  & 2921\\
Madison    & 2921\\
Monroe     & 2921\\
Adams      & 1460\\
Jackson    & 2921\\
VanBuren   & 1460\\
Harrison   & 31\\
Tyler      & 1427\\
Polk       & 1460\\
Taylor     & 491\\
Filmore    & 967\\
Pierce     & 1460\\
Buchanan   & 1460\\
Lincoln    & 1503\\
Johnson    & 1418\\
Grant      & 2921\\
Hayes      & 1460\\
Garfield   & 199\\
Arthur     & 1260\\
Cleveland  & 1460\\
Harrison   & 1460\\
Cleveland  & 1460\\
McKinley   & 1655\\
Roosevelt  & 2727\\
Taft       & 1460\\
Wilson     & 2921\\
Harding    & 881\\
Coolidge   & 2039\\
Hoover     & 1460\\
Roosevelt  & 4452\\
Truman     & 2810\\
Eisenhower & 2922\\
Kennedy    & 1036\\
Johnson    & 1886\\
Nixon      & 2027\\
Ford       & 895\\
Carter     & 1461\\
Reagan     & 2922\\
Bush       & 1461\\
Clinton    & 2922\\
Bush       & 1110\\
\bottomrule\\
\end{tabular}
\caption{Presidents of the United States of America: days in office.}
\label{tab:presidents}
\end{table}
 
\clearpage
\section{Hierarchical logistic regression model}
\label{app:hier_log_reg}

Hierarchical logistic regression models classification tasks in an intuitive
way. We study three variants of \citet{gelman2006data}'s model for the 1988 CBS
United States election survey dataset. 

\paragraph{First model: no age or education.}
The likelihood of voting Republican is
\begin{align*}
  \Pr(y_n=1)
  &=
  \text{sigmoid}
  \bigg(
  \beta^\text{female}\cdot\text{female}_n
  + \beta^\text{black}\cdot\text{black}_n
  + \alpha^\text{state}_{s[n]}
  \bigg),
\end{align*}
with priors
\begin{align*}
  \alpha^\text{state}_j
  &\sim
  \cN
  \left(
  \mu_\text{state}
  \,,\, \sigma_\text{state}
  \right)\\
  \mu_\text{state} &\sim \cN(0,10)\\
  \sigma_\text{state} &\sim \cN(0,10)\\
  \mbbeta &\sim \cN(0,1).
\end{align*}

\paragraph{Second model: with age.}
The likelihood of voting Republican is
\begin{align*}
  \Pr(y_n=1)
  &=
  \text{sigmoid}
  \bigg(
  \beta^\text{female}\cdot\text{female}_n
  + \beta^\text{black}\cdot\text{black}_n
  + \alpha^\text{state}_{s[n]}
  + \alpha^\text{age}_{a[n]}
  \bigg),
\end{align*}
with priors
\begin{align*}
  \alpha^\text{state}_j
  &\sim
  \cN
  \left(
  \mu_\text{state}
  \,,\, \sigma_\text{state}
  \right)\\
  \mu_\text{state} &\sim \cN(0,10)\\
  \sigma_\text{state} &\sim \cN(0,10)\\
  \alpha^\text{age}_j
  &\sim
  \cN
  \left(
  \mu_\text{age}
  \,,\, \sigma_\text{age}
  \right)\\
  \mu_\text{age} &\sim \cN(0,10)\\
  \sigma_\text{age} &\sim \cN(0,10)\\  
  \mbbeta &\sim \cN(0,1).
\end{align*}

\paragraph{Third model: with education.}
The likelihood of voting Republican is
\begin{align*}
  \Pr(y_n=1)
  &=
  \text{sigmoid}
  \bigg(
  \beta^\text{female}\cdot\text{female}_n
  + \beta^\text{black}\cdot\text{black}_n
  + \alpha^\text{state}_{s[n]}
  + \alpha^\text{edu}_{a[n]}
  \bigg),
\end{align*}
with priors
\begin{align*}
  \alpha^\text{state}_j
  &\sim
  \cN
  \left(
  \mu_\text{state}
  \,,\, \sigma_\text{state}
  \right)\\
  \mu_\text{state} &\sim \cN(0,10)\\
  \sigma_\text{state} &\sim \cN(0,10)\\
  \alpha^\text{edu}_j
  &\sim
  \cN
  \left(
  \mu_\text{edu}
  \,,\, \sigma_\text{edu}
  \right)\\
  \mu_\text{edu} &\sim \cN(0,10)\\
  \sigma_\text{edu} &\sim \cN(0,10)\\  
  \mbbeta &\sim \cN(0,1).
\end{align*}

\clearpage
\section{Hierarchical Poisson factorization model}
\label{app:hpf}

Hierarchical Poisson factorization models a matrix of counts as a
low-dimensional inner product. \citep{gopalan2015scalable} present the model in
detail along with an efficient variational inference algorithm for inference. We
summarize the model below. 

\paragraph{Model.} Consider a $U \times I$ dataset, with non-negative integer
elements $x_{ui}$. It helps to think of $u$ as an index over ``users'' and $i$
as an index over ``items''.

The likelihood for each measurements is
\begin{align*}
  p(x_{ui})
  &=
  \text{Poisson}(x_{ui}\prm \mbtheta_u^\top \mbbeta_i)
\end{align*}
where $\mbtheta$ is a $U\times K$ matrix of non-negative real-valued latent
variables; it represents ``user preferences''. Similarly, $\mbbeta$ is a 
$K\times I$ matrix of non-negative real-valued latent variables; it represents
``item attributes''.

The priors for these variables are
\begin{align*}
  p(\theta_{uk}) 
  &= 
  \Gam(\theta_{uk} \prm a, \xi_u)\\
  p(\beta_{ki}) 
  &= 
  \Gam(\beta_{ki} \prm c, \eta_i)
\end{align*}
where $\mbxi$ is a $U$ vector of non-negative real-valued latent variables; it
represents ``user activity''. Similarly, $\mbeta$ is a $I$ vector of
non-negative real-valued latent variables; it represents ``item popularity''.

The prior for these hierarchical latent variable are
\begin{align*}
  p(\xi_u)
  &=
  \Gam(\xi_u\prm a^\prime,\nicefrac{a^\prime}{b^\prime})\\
  p(\eta_i)
  &=
  \Gam(\eta_i\prm c^\prime,\nicefrac{c^\prime}{d^\prime}).
\end{align*}

\paragraph{Parameters for IRi analysis.} The parameters we used were
\begin{align*}
  K &= 20\\
  a &= \nicefrac{0.3}{\sqrt{K}}\\
  c &= \nicefrac{0.3}{\sqrt{K}}\\
  a^\prime &= 1.5\\
  b^\prime &= 0.3\\
  c^\prime &= 1.5\\
  d^\prime &= 0.3.
\end{align*}

\textbf{Inference.} We used the coordinate ascent variational inference
algorithm provided by the authors of \citep{gopalan2015scalable}.

\clearpage
\section{Population genetics data and model}
\label{app:admixture}

Population genetics studies ancestral trends of genomic mutations.
Consider $N$ individuals and $L$ locations on the genome. For each
location, we measure whether each individual reveals a mutation. This gives an 
($N \times L$) dataset $\mbx$ where $x_{nl} \in {0,1,2,3}$. (We assume two
specific forms of mutation; 3 encodes a missing observation.)

\textbf{Model.}
\citet{pritchard2000inference} propose a probabilistic model for this kind of
data. Represent $K$ ancestral populations with a latent variable $\mbphi$. 
This is a $(K \times L)$ matrix of mutation probabilities. Place a
beta prior for each probability. Each individual mixes these populations. Denote
this with another latent variable $\mbtheta$. This is a $(N\times K)$ matrix of 
mixture proportions. Place a Dirichlet prior for each individual.
The likelihood of each mutation is a $K$-mixture of categorical distributions. 

\textbf{Data.}
We study a subset of the Hapmap 3 dataset from \citep{mimno2015posterior}. This
includes $N=324$ individuals and $L=13\,928$ locations on the genome. Four
geographic regions are represented: 49 \textsc{asw}, 112 \textsc{ceu}, 50 
\textsc{mex}, and 113 \textsc{yri}. \citet{mimno2015posterior} expect this data
to exhibit at most $K=3$ ancestral populations; the full Hapmap 3 dataset
exhibits $K=6$ populations. 

In more detail, this data studies \glspl{SNP}. The data is pre-processed from
the raw genetic observations such that:
\begin{itemize}
  \item non-\glspl{SNP} are removed (i.e.~genes with more than one location
  changed),
  \item \glspl{SNP} with low entropy compared to the
  dominant mutation at each location are removed,
  \item \glspl{SNP} that are too close to each other on the genome are removed.
\end{itemize}

\textbf{Inference.} 
We use the fastSTRUCTURE software suite \citep{raj2014faststructure} to perform
variational inference. We use the default parameters and only specify $K=3$.

\end{document}